\documentclass[11pt]{article}
\usepackage[a4paper,margin=25mm]{geometry}
\usepackage[T1]{fontenc}
\usepackage{lmodern}
\usepackage{amsmath,amssymb,graphicx,booktabs,array,microtype}
\usepackage[font=small,labelfont=bf]{caption}
\usepackage[hidelinks]{hyperref}
\usepackage[section]{placeins}
\DeclareMathOperator{\LN}{LN}
\DeclareMathOperator{\SiLU}{SiLU}
\DeclareMathOperator{\MSE}{MSE}
\DeclareMathOperator{\clip}{clip}
\DeclareMathOperator{\sg}{sg}
\DeclareMathOperator{\softmax}{softmax}
\title{Beyond Emotion Prompts: Fine-Grained Text-to-Image Generation Driven by Valence--Arousal--Dominance}
\author{Minglang Li \quad Yueyue Fang \quad Xieping Gao\thanks{Corresponding author. Email: \href{mailto:202510290587@hunnu.edu.cn}{202510290587@hunnu.edu.cn}.}\\[0.5em]
\small College of Information Science and Engineering\\
\small Hunan Normal University, Changsha, China}

\date{}
\begin{document}
\maketitle
\begin{abstract}
Although text-to-image models can accurately depict subjects and scenes, creators still struggle to specify the fine-grained emotions an image should convey without rewriting its content description. Natural language can suggest emotions, but it offers no control scale with stable meanings and ordered intensities. We propose EMOTRANS, which transforms psychologically grounded valence--arousal--dominance (VAD) coordinates into generation conditions that are independent of the content text and modulated across denoising stages, making emotional style a finely adjustable creative variable. To support this goal, we construct EMOVAD, an art-painting dataset that pairs objective content descriptions with separately collected emotional ratings from multiple annotators. We also coordinate emotional expression and content preservation through dual-branch training with a shared model. Objective and human evaluations show that the framework improves the accuracy of three-dimensional emotion control and produces perceptible, orderable continuous changes while maintaining competitive text alignment and image quality. This work provides a practical emotion-driven approach to image generation that extends objective content depiction to fine-grained emotional adjustment.
\end{abstract}
\section{Introduction}
Text-to-image generation is evolving from the visual reproduction of language descriptions into an expressive tool for artistic creation and interactive design; work on media narrative modeling has also expanded how visual content can be organized~\cite{r1}. Latent diffusion and Transformer-based generative backbones continue to improve image detail, semantic alignment, and generation efficiency~\cite{r2,r3,r4}, while recent studies further extend high-resolution synthesis and multimodal conditioning~\cite{r5,r6}. However, an image's objective semantic content is not equivalent to its emotional meaning. The same subjects and scenes can convey very different emotional atmospheres through color, lighting, posture, and spatial relationships. For creative workflows involving repeated comparison and refinement, producing images with the correct content is insufficient. Enabling emotional expression to be explicitly specified and continuously adjusted while preserving content constraints is therefore an important step toward more precise text-to-image creation.

\begin{figure}[tbp]
\centering
\includegraphics[width=1\linewidth]{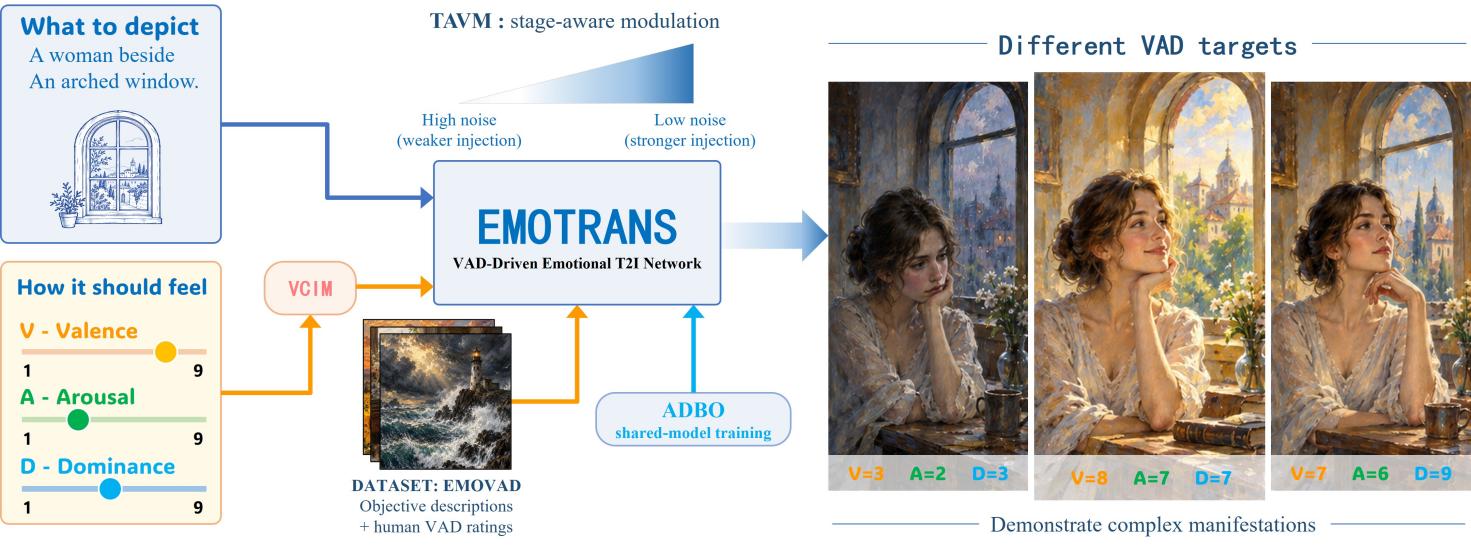}
\caption{The central idea of EMOTRANS. VAD coordinates specify emotional expression; VCIM establishes an independent emotion pathway, TAVM modulates its strength across denoising stages, and ADBO coordinates emotion control with content preservation during training. EMOVAD provides separately paired supervision. The examples on the right illustrate fine-grained emotional differences for the same scene under different VAD targets.}
\label{fig:concept}
\end{figure}

Research addressing this problem has progressed from general conditional control toward emotion-aware analysis and generation. ControlNet and subsequent unified control models incorporate structural conditions and reference images into generation~\cite{r7,r8}, while style-transfer studies examine the tension between appearance control and content leakage~\cite{r9,r10}. For emotional generation, EmoGen uses emotion-related visual factors~\cite{r11}, EmotiCrafter introduces continuous V--A conditions~\cite{r12}, and EmoStyle and CoEmoGen explore emotion-driven artistic stylization and psychologically inspired hierarchical emotion modeling, respectively~\cite{r13,r14}. These advances indicate that emotion control cannot be reduced to adding adjectives to prompts. Nevertheless, the ability of emotional conditions to influence an image does not establish that expression along each dimension is clearly differentiated. When the target emotion changes, a model must still reconcile conditional expression with content preservation.

Addressing this problem requires more than reusing conditioning strategies for visible attributes such as object layout. Emotion-oriented text-to-image generation also needs emotion psychology to define the meaning and scale of its control variables. Dimensional theories describe emotional differences through valence, arousal, and dominance, corresponding to pleasantness, activation, and feelings of control or being controlled~\cite{r15,r16}. Measurement tools such as the Self-Assessment Manikin (SAM) provide operational rating procedures for these dimensions~\cite{r17}. Compared with discrete emotion words, VAD coordinates permit intensity variation within an emotion and allow differences in perceived control to be specified separately for similar V--A states. Accordingly, objective image content and viewers' emotional responses in Figure~\ref{fig:concept} should be treated as inputs with distinct but coordinated roles: text constrains content, while psychological coordinates specify emotional direction and degree. Interpretability here derives from the explicit meanings of the input dimensions and their traceable network pathways, rather than an assumed fixed correspondence between an emotion and a particular color or posture.

We propose EMOTRANS, a diffusion Transformer (DiT) framework that implements this dual-input design through representation, timing, and optimization. The VAD Conditioning and Injection Module (VCIM) encodes low-dimensional VAD values into global emotion tokens and injects them into image features through independent cross-attention residuals, reducing competition with content text within the same conditioning channel. Timestep-Adaptive VAD Modulation (TAVM) uses the signal-retention coefficient of the noise schedule to weaken emotional intervention at high noise levels and strengthen emotional refinement at low noise levels. The Adaptive Dual-Branch Objective (ADBO) jointly trains a VAD-conditioned branch and a text-only branch of the same model, reusing the temporal weights to coordinate their denoising supervision. Unlike general dynamic condition fusion~\cite{r18}, our schedule specifically modulates emotion-driven residual connections and aligns injection strength with the dual-branch objective.

To provide independent supervision, we construct EMOVAD by separately pairing art paintings, objective descriptions, and human VAD ratings. We select paintings as the first controlled validation domain, without assuming that paintings represent the entire visual world. Painting actively organizes emotional experience through composition, color, lighting, and brushwork, and prior work has treated artistic content and style as important vehicles for emotional expression~\cite{r13,r19}. In open-domain photographs, emotional interpretation is often intertwined with event context, viewer background, and social circumstances~\cite{r20}. We therefore use painting as a focused setting in which to examine whether continuous VAD axes produce stable, distinguishable visual responses under explicit medium and content constraints. This scope supports validation of the control mechanism; it does not imply that the findings already generalize to all visual domains.

Our main contributions are as follows:
\begin{itemize}
\item We introduce EMOTRANS and design VCIM for independent VAD representation and layer-wise injection. This enables continuous emotional adjustment, including dominance, under textual content constraints, with explicit psychological meanings and network pathways.
\item We design TAVM and ADBO to couple emotional residual injection with dual-branch denoising supervision through shared temporal weights, coordinating fine-grained emotional expression and content preservation across generation stages.
\item We construct EMOVAD, which separately pairs art paintings, objective descriptions, and VAD ratings from multiple annotators, providing supervision for three-dimensional emotion control independently of explicit emotional wording.
\end{itemize}

\section{Related Work}
\subsection{Text-to-Image Generation and Conditional Control}
In text-to-image generation without additional conditions such as layouts or reference images, early research focused on translating language semantics into high-fidelity visual content. DDPM established the basis of iterative denoising~\cite{r21}; latent diffusion reduced the cost of high-resolution synthesis~\cite{r2}; and DiT and PixArt-$\alpha$ advanced scalable Transformer backbones and efficient training~\cite{r3,r4}. SANA, SANA 1.5, and Lumina-Image 2.0 further improve efficiency and semantic adherence through linear attention, scaling of training and inference compute, and unified text--image representations~\cite{r5,r22,r23}. Surveys also identify the joint development of model architectures, language representations, and data quality as a major route to stronger generation~\cite{r24}. These advances provide reusable generative foundations, but better image quality alone does not establish an explicit scale for emotion control.

When creative requirements exceed what text can describe precisely, additional conditions are introduced into denoising. ControlNet supplements textual constraints with structural conditions~\cite{r7}; OmniGen integrates multiple image-generation tasks within a unified framework~\cite{r6}; and OminiControl supports spatially aligned and unaligned visual conditions with a lightweight design~\cite{r8}. Conditional control has thus moved from task-specific interfaces toward more general input organization, raising the question of how multiple conditions should be coordinated~\cite{r25}. DynFusion dynamically fuses conditions according to the timestep, task, and injection location~\cite{r18}. This development suggests that effective control depends not only on additional information but also on where and how strongly it enters generation. Emotional conditions likewise require mechanisms suited to their semantics.

Among non-emotional conditions, UNO addresses reference consistency from single to multiple subjects~\cite{r26}, while Compass Control explicitly controls object orientation~\cite{r27}. These approaches primarily constrain visible identity or geometry. Visual style research instead addresses the entanglement of appearance and content: StyleStudio selectively transfers style elements~\cite{r9}, StyleKeeper suppresses content leakage from style references~\cite{r10}, and DuoLoRA coordinates content--style personalization~\cite{r28}. CoTyle compresses appearance style into discrete codes~\cite{r29}, further simplifying the control interface. These studies inform the separation of conditioning pathways and content preservation. However, reference appearance, geometric parameters, and discrete style indices do not inherently carry the psychological meanings or ordering of emotional intensity, and therefore cannot directly replace VAD coordinates.

Emotion-conditioned generation extends control from visible attributes to viewers' feelings. RePrompt improves emotional expression by editing prompts~\cite{r30}; EmoGen introduces emotion-related visual factors into diffusion generation~\cite{r11}; and EmotiCrafter uses valence and arousal as continuous conditions~\cite{r12}. More recently, EmoStyle models emotional style for artistic generation~\cite{r13}, and CoEmoGen combines emotional content descriptions with hierarchical adaptation to improve semantic coherence~\cite{r14}. In the adjacent area of emotional editing, EmoEdit, EmoEditor, and AIEdiT investigate changing the emotional expression of existing images~\cite{r31,r32,r33}. Together, these works advance emotion from a label to a manipulable condition. We focus further on continuous coordinates that include dominance and their coordination with separate content and emotion pathways, stage-dependent modulation, and shared dual-branch optimization. We do not claim to introduce emotional generation or dynamic conditional control as entirely new problems.

\subsection{VAD Representations and Psychologically Driven Generation}
Dimensional emotion research describes relationships among emotions using a small number of psychologically meaningful coordinates. Russell and Mehrabian's three-factor study provides empirical support for pleasure, arousal, and dominance~\cite{r15}, while Russell's circumplex model characterizes the two-dimensional organization of valence and arousal~\cite{r16}. These accounts are related, but the two-dimensional model is not equivalent to the full VAD representation. SAM subsequently translated all three dimensions into intuitive self-assessment procedures~\cite{r17}, and large-scale lexical norms provided empirical dimensional ratings~\cite{r34}. This research supports treating emotion as ordered rather than exclusively discrete. It also highlights the distinction between psychological scales and visual realizations: the same coordinates can be expressed through multiple visual cues and are affected by context and individual differences.

The integration of psychology and visual AI has expanded from emotion recognition to supervision and intervention in generation. A review of affective image analysis examines the joint roles of content, viewers, and context~\cite{r20}; ArtEmis links art images to emotional language~\cite{r19}; and EmoSet provides emotional annotations with rich visual attributes~\cite{r35}. More recently, EmoArt connects emotion categories in artistic data with A--V information~\cite{r36}, providing more detailed resources for emotion-aware generation. Work on affective filters based on generative priors explores how textual emotions can become visual style changes~\cite{r37}. These studies bring psychological concepts into AI through annotations, conditional representations, and visual operations. However, emotion labels in a dataset do not automatically ensure independent, continuous, and stable model responses along every psychological dimension.

We therefore establish an operational connection between psychological representations and emotional generation: VAD coordinates including dominance define the control targets, EMOVAD separates content descriptions from emotional ratings, and EMOTRANS provides an independent pathway for numerical conditions. The aim is to make changes along clearly defined dimensions correspond to traceable network interventions while preserving textual content as far as possible. VCIM, TAVM, and ADBO address representational mismatch, stage-dependent differences, and coordination of training objectives, respectively, motivating the method below.

\section{Methodology}
\subsection{The EMOVAD Dataset}
\begin{figure}[tbp]
\centering
\includegraphics[width=1\linewidth]{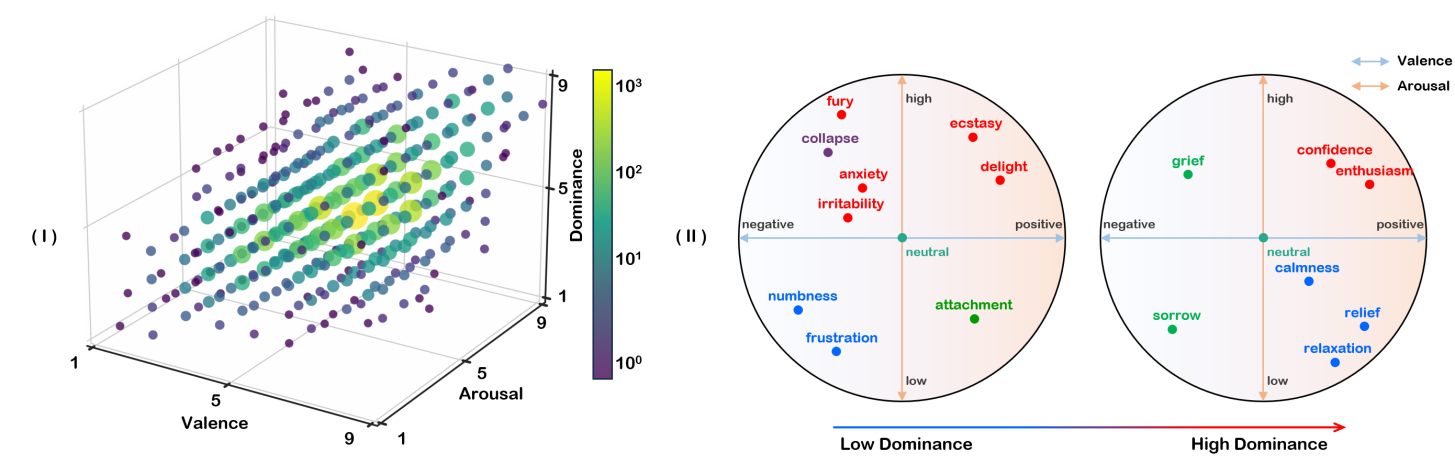}
\caption{VAD distribution and emotional-space illustration for EMOVAD. (I) Empirical sample frequencies in the three-dimensional rating space. (II) Relative positions of representative emotions at different dominance levels, illustrating the meaning of the three-dimensional representation rather than defining fixed coordinates for emotion categories.}
\label{fig:dataset}
\end{figure}

Independent VAD control first requires separating supervision of image content from supervision of emotion. If descriptions explicitly include the target emotion, a model may rely on linguistic cues and ignore numerical conditions, making it difficult to determine whether VAD actually drives visual changes. EMOVAD therefore separately pairs art paintings, objective content descriptions, and human three-dimensional ratings, without including emotion categories or VAD values in the content text. Its annotations follow the dimensional representation discussed in Section~2.2. Figure~\ref{fig:dataset}(II) illustrates relationships among emotional states in the V--A plane at different dominance levels. This schematic emphasizes that D adds a distinction between states with similar V--A values; it does not assign a unique coordinate to each emotion category.

Even psychologically grounded VAD coordinates can be bypassed by a model if rating noise is high or the text directly reveals emotional targets. We therefore use human ratings, review by participants from different backgrounds, and objective content descriptions to establish relatively independent text and VAD annotations. Artistic paintings provide comparatively concentrated emotional cues; mixing media and semantics across a fully open domain would introduce substantial attribution noise. We thus validate continuous VAD control in a controlled but diverse painting domain.

Specifically, EMOVAD selects 14,637 representative paintings from WikiArtimage. Annotators with a psychology background manually rate each image on all three VAD dimensions. Each dimension uses a 1--9 scale to preserve ordered variation from low to high. As shown in Figure~\ref{fig:dataset}(I), the samples neither cluster at a single VAD point nor collapse onto a single plane, but cover an ordered lattice along all three axes. The empirical frequencies indicated by color are higher in the middle of the rating space and decrease toward extreme boundaries, consistent with moderate judgments being more common than extreme judgments on bounded psychological scales. EMOVAD therefore combines a stable central sample density with conditional coverage extending in different emotional directions, supporting learning of continuous changes from neutral to different intensity regions.

To prevent explicit emotional leakage through text, descriptions are kept objective and neutral. They record only observable subjects, objects, actions, scenes, and spatial relationships, excluding explicit emotion categories, viewers' feelings, and evaluative wording. All descriptions are manually annotated and verified, yielding a text corpus of 1,661,288 words. The final dataset contains 73,185 human annotation records, corresponding to five rating records per image. To reduce biases associated with a single population or data processing, 30 participants of different ages, disciplinary backgrounds, and levels of psychological knowledge comprehensively review and clean the annotations.

\subsection{EMOTRANS}
\begin{figure}[tbp]
\centering
\includegraphics[width=1\linewidth]{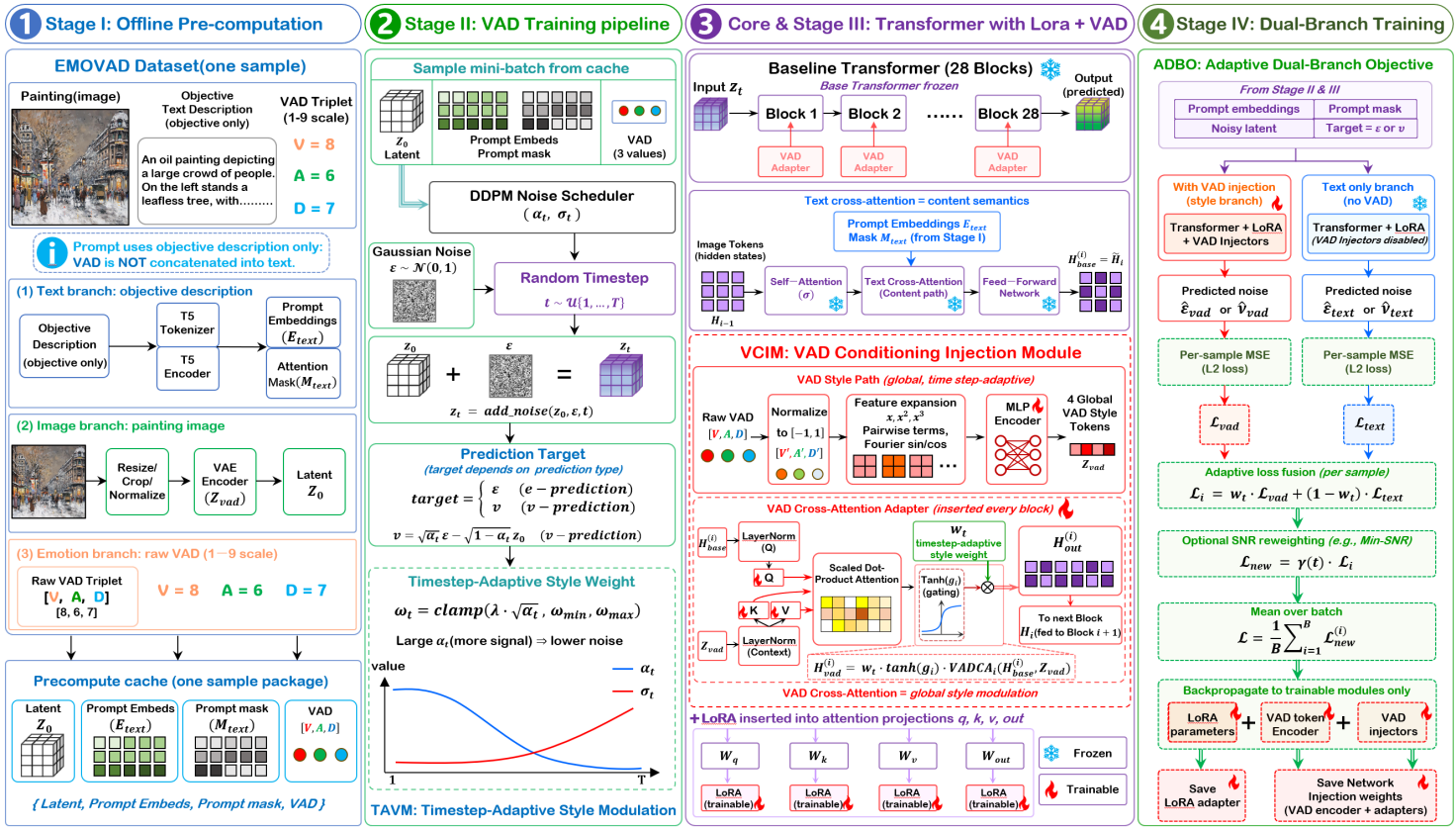}
\caption{Overall architecture of EMOTRANS. Separate pathways process textual content and VAD emotion. VCIM encodes and injects emotional conditions at each layer; TAVM modulates timestep weights; and ADBO coordinates VAD-conditioned and text-conditioned supervision through two forward passes of a shared model. The PixArt-$\alpha$ base weights, T5, and VAE are frozen. Only LoRA, the VAD encoder, and layer-specific adapters are updated.}
\label{fig:architecture}
\end{figure}

Figure~\ref{fig:concept} presents separate content and emotion inputs, and Figure~\ref{fig:architecture} shows their implementation in a diffusion Transformer. EMOTRANS addresses three questions: how low-dimensional VAD becomes a queryable visual condition, when emotional residuals should be strengthened, and how training preserves content-generation ability without VAD. VCIM, TAVM, and ADBO address these questions, respectively, and form a coordinated control mechanism through shared emotional representations and temporal weights.

The model builds on pretrained PixArt-$\alpha$~\cite{r4}. It retains the original text cross-attention for subject, scene, and relational semantics while introducing separate VAD cross-attention for global emotional modulation. Frozen VAE and T5 encoders produce image latents, text embeddings, and masks from images and objective text; raw VAD ratings enter VCIM independently. The middle of Figure~\ref{fig:architecture} and Figure~\ref{fig:vcim} show construction and layer-wise residual injection of VAD tokens. Figure~\ref{fig:tavm} illustrates stage-dependent injection modulation, and the right of Figure~\ref{fig:architecture} shows shared-model dual-branch training. Base-model weights remain frozen; only LoRA updates~\cite{r38}, the VAD encoder, and layer-specific adapters are trained. Separating the pathways reduces direct competition, but does not assume complete independence of content and emotion in visual space.

\subsubsection{VCIM: VAD Conditioning and Injection Module}
\begin{figure}[tbp]
\centering
\includegraphics[width=0.78\linewidth]{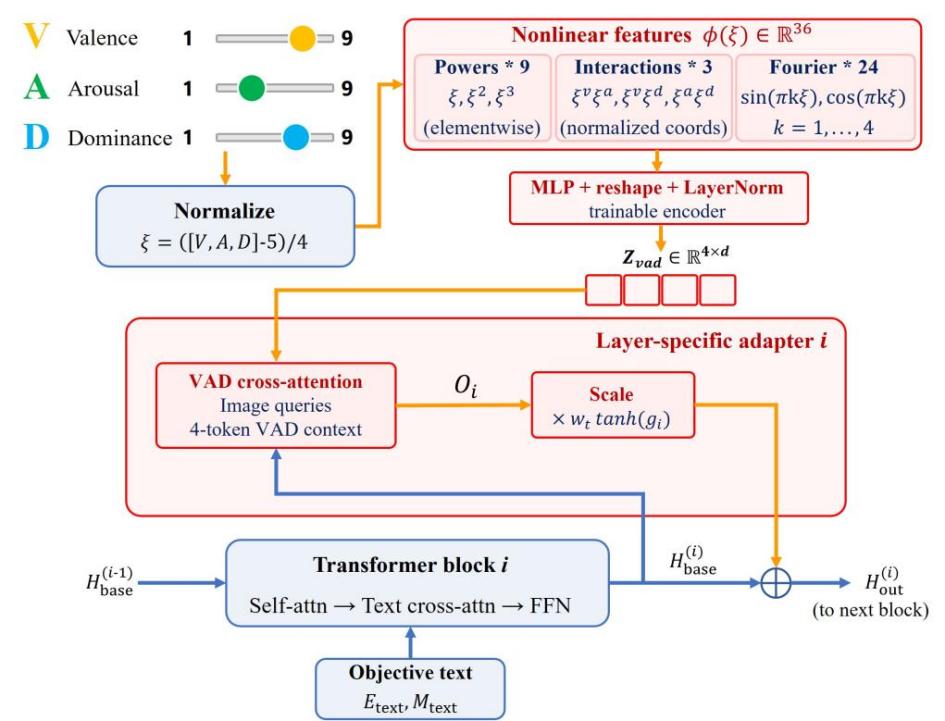}
\caption{Condition encoding and layer-wise injection in VCIM. Normalized VAD is expanded into 36 features and mapped to four global emotion tokens. At each layer, image features query the VAD context to generate a residual, which is scaled by the temporal weight and a layer-specific gate before being added to the content pathway.}
\label{fig:vcim}
\end{figure}

Three VAD scalars cannot directly meet the representational requirements of high-dimensional image tokens, while rewriting them as emotion words would make control depend on textual wording again. VCIM therefore separates how emotion is represented from where it influences images. As shown in Figure~\ref{fig:vcim}, nonlinear features and a trainable encoder form a compact emotional context. Image features produced by the content pathway query this context, and a gated residual performs fusion. The blue text pathway continues to constrain content, while the orange pathway provides adjustable global emotional information.

To retain within-axis variation and explicitly represent interactions between axes, VCIM applies nonlinear feature expansion before the MLP rather than a single linear projection of the three scores. Raw scores are first centered and scaled:
\begin{equation}
\boldsymbol{\xi}=(\xi_V,\xi_A,\xi_D)=\frac{(V,A,D)-5}{4}\in[-1,1]^3.
\end{equation}
Here, $\boldsymbol{\xi}$ is the normalized VAD vector, with components $\xi_V$, $\xi_A$, and $\xi_D$. Figure~\ref{fig:vcim} divides the 36 features into three groups: nine first-, second-, and third-order powers of the three components; three pairwise products for V--A, V--D, and A--D; and 24 sine and cosine terms over four frequencies for each component. These groups represent within-axis nonlinearity, between-axis interactions, and multifrequency variation. They increase encoding capacity without assigning a particular visual emotion to any feature. A two-layer SiLU MLP and LayerNorm then transform these features into four global VAD tokens:
\begin{align}
h_1 &= \LN_1\bigl(\SiLU(W_1\phi(\boldsymbol{\xi})+b_1)\bigr),\\
h_2 &= \LN_2\bigl(\SiLU(W_2h_1+b_2)\bigr),\\
Z_{\mathrm{vad}} &= \LN_3\bigl(\operatorname{reshape}(W_3h_2+b_3,B,K,d)\bigr),\qquad K=4.
\end{align}
The function $\phi$ constructs the explicit 36-dimensional feature vector. The $W_j$ and $b_j$ are learnable projection matrices and biases, and $h_1,h_2$ are the two hidden representations. $\LN_j$ denotes LayerNorm along the final dimension; SiLU denotes the sigmoid linear unit; and reshape changes tensor dimensions without reordering elements. $B$ is the batch size, $K$ is the number of VAD style tokens, and $d$ is the PixArt hidden dimension. The default hidden-layer width is 256. A single VAD encoder is shared across 28 Transformer blocks, so each sample requires only one compact global emotional context. The four tokens jointly encode three-dimensional VAD rather than corresponding individually to the psychological axes. The context is shared across spatial positions, but image queries and attention responses can vary by location.

After constructing VAD tokens, we must determine how they affect image features. Concatenating them directly with text tokens would return emotion to the content channel, while injection at only one layer would inadequately accommodate representations at different depths. We therefore introduce independent VAD cross-attention at every layer, providing depth-dependent global modulation while retaining the original text pathway. Let the original content pathway of the $i$th PixArt block be
\begin{equation}
H_{\mathrm{base}}^{(i)}=\mathcal{F}_i\bigl(H_{\mathrm{out}}^{(i-1)},E_{\mathrm{text}},M_{\mathrm{text}},t\bigr).
\end{equation}
Here, $i$ indexes Transformer blocks. $H_{\mathrm{out}}^{(i-1)}$ and $H_{\mathrm{base}}^{(i)}$ are the input state and content-pathway output of block $i$. The mapping $\mathcal{F}_i$ combines frozen original PixArt weights with trainable LoRA updates. $E_{\mathrm{text}}$ denotes the text embeddings, $N_{\mathrm{text}}$ the number of text tokens, $M_{\mathrm{text}}$ the text-token mask, and $t$ the diffusion timestep. For attention head $h$,
\begin{align}
Q_i^h &= W_i^{Q,h}\LN_q(H_{\mathrm{base}}^{(i)}),\\
K_i^h &= W_i^{K,h}\LN_k(Z_{\mathrm{vad}}),\qquad
U_i^h = W_i^{U,h}\LN_k(Z_{\mathrm{vad}}),\\
O_i &= W_i^O\operatorname{Concat}_{h=1}^{H}\left[
\softmax\left(\frac{Q_i^h(K_i^h)^\top}{\sqrt{d_h}}\right)U_i^h\right].
\end{align}
The image state $H_{\mathrm{base}}^{(i)}$ and global VAD context $Z_{\mathrm{vad}}$ are defined above. $H$ is the number of attention heads and $d_h$ is the channel width per head. $Q_i^h$ contains image queries; $K_i^h$ and $U_i^h$ are VAD keys and values. $W_i^{Q,h}$, $W_i^{K,h}$, $W_i^{U,h}$, and $W_i^O$ are the learnable query, key, value, and output projections of adapter $i$. $\LN_q$ and $\LN_k$ normalize queries and the VAD context, respectively, and $1/\sqrt{d_h}$ is the scaling factor. $O_i$ is the VAD adapter output; $U_i^h$ denotes the value projection. Although the encoder is shared, all 28 adapters are independent, allowing different depths to learn different emotional responses. The gated residual is
\begin{equation}
H_{\mathrm{vad}}^{(i)}=\widehat{w}_t\tanh(g_i)O_i,\qquad
H_{\mathrm{out}}^{(i)}=H_{\mathrm{base}}^{(i)}+H_{\mathrm{vad}}^{(i)}.
\end{equation}
Here, $g_i$ is a learnable layer-specific scalar gate, restricted to $(-1,1)$ by $\tanh$. The timestep weight $\widehat{w}_t$ comes from TAVM and is broadcast across token and channel dimensions. The injected residual and content state have the same shape, and their sum gives the block output.

\subsubsection{TAVM: Timestep-Adaptive VAD Modulation}
\begin{figure}[tbp]
\centering
\includegraphics[width=0.83\linewidth]{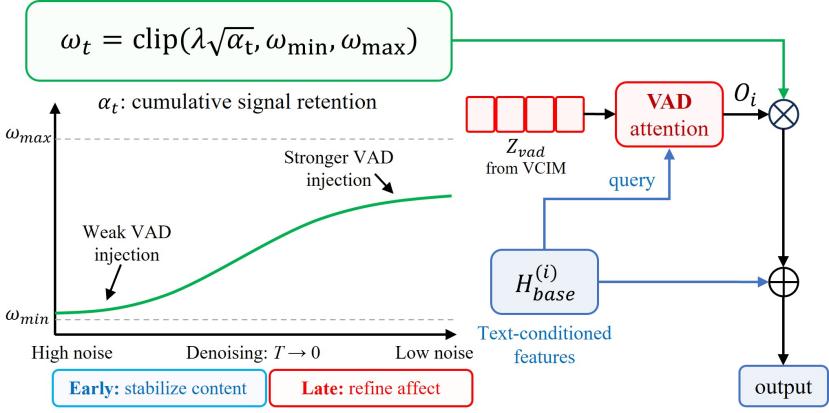}
\caption{Timestep-adaptive modulation in TAVM. Along denoising from high to low noise, the bounded weight increases from weak to strong.}
\label{fig:tavm}
\end{figure}

Fixed-strength VAD injection ignores differences in signal reliability across denoising stages. Strong emotional intervention can increase content drift before subjects and layouts stabilize, whereas stronger modulation at low noise levels is better suited to refining atmosphere and expressive cues. TAVM adopts the stage-adaptive strategy in Figure~\ref{fig:tavm}, directly using the scheduler's cumulative signal-retention coefficient without an auxiliary prediction network. Along the denoising direction from $T$ to $0$, emotional weights increase from weak to strong. This is a design prior for coordinating content and emotion, rather than a claim that emotional generation occurs exclusively within a separate stage.

Let $\beta_s$ be the single-step noise variance and $\alpha_t$ the cumulative signal-retention coefficient. The forward noising process is
\begin{equation}
\alpha_t=\prod_{s=1}^{t}(1-\beta_s),\quad
\sigma_t=\sqrt{1-\alpha_t},\quad
z_t=\sqrt{\alpha_t}\,z_0+\sigma_t\epsilon,\quad
\epsilon\sim\mathcal{N}(0,I).
\end{equation}
The index $s$ runs over the noise schedule, and $\beta_s$ is the noise variance at step $s$. Thus $\alpha_t$ denotes cumulative signal retention through step $t$, while $\sqrt{\alpha_t}$ is the clean-latent amplitude coefficient. $\sigma_t$ is the cumulative noise standard deviation. $z_0$ and $z_t$ are clean and noisy VAE latents, and $\epsilon$ is independent standard Gaussian noise of the same shape. $\mathcal{N}(0,I)$ denotes a Gaussian distribution with zero mean and identity covariance, and $\prod$ denotes a product. We define the bounded modulation weight as
\begin{equation}
w_t=\clip\left(\lambda\sqrt{\alpha_t},w_{\min},w_{\max}\right).
\end{equation}
Here, $\lambda$ scales the style weight, $w_{\min}$ and $w_{\max}$ are the minimum and maximum permitted strengths, and $\clip$ truncates values to this interval. $w_t$ is therefore the bounded base VAD weight at timestep $t$. With VAD condition dropout during training, the base weight is multiplied by a per-sample condition-retention mask to obtain the effective weight $\widehat{w}_t$. Without dropout, these weights coincide; injection is disabled in the text-only pathway. Within the weight range used in this work, the same effective weight controls both residual modulation and dual-branch loss mixing.

As shown on the right of Figure~\ref{fig:tavm}, TAVM scales the emotional residual computed by VCIM, rather than textual conditions or base content features. The full injection also includes the layer-specific gate in Figure~\ref{fig:vcim}. This mechanism requires no additional layout labels, style labels, or auxiliary predictor. Reusing the effective weight in ADBO synchronizes feature intervention with the emphasis of the training objective. Denoising supervision uses only the first four output channels, with the target selected according to the scheduler's prediction type:
\begin{equation}
y_t=\begin{cases}
\epsilon, & \epsilon\text{-prediction},\\
v_t=\sqrt{\alpha_t}\,\epsilon-\sigma_tz_0, & v\text{-prediction}.
\end{cases}
\end{equation}
Here, $y_t$ is the denoising target. For $\epsilon$-prediction it equals the added noise; for $v$-prediction it equals the velocity target $v_t$. The latter has the same shape as $z_0$ and is determined by $\alpha_t$, $\sigma_t$, $\epsilon$, and $z_0$. The noise scheduler selects the prediction type.

\subsubsection{ADBO: Adaptive Dual-Branch Objective}
An independent emotional pathway and temporal modulation alone do not guarantee content preservation: a model could reduce emotion-conditioned denoising error by changing subjects or scenes. ADBO therefore adds a text-only reference pathway with VAD disabled, ensuring that parameter updates are also constrained by text-conditioned generation. As shown on the right of Figure~\ref{fig:architecture}, both branches use the same noisy sample, timestep, and text conditions, and perform two forward passes through one model with fully shared parameters $\Theta$. They are neither independent generators nor a frozen teacher--student pair:
\begin{align}
\widehat{y}_{\mathrm{vad}}^{(n)}
&=\pi_4 f_\Theta\bigl(z_t^{(n)},t_n,E_{\mathrm{text}}^{(n)},M_{\mathrm{text}}^{(n)},Z_{\mathrm{vad}}^{(n)};\widehat{w}_{t,n}\bigr),\\
\widehat{y}_{\mathrm{text}}^{(n)}
&=\pi_4 f_\Theta\bigl(z_t^{(n)},t_n,E_{\mathrm{text}}^{(n)},M_{\mathrm{text}}^{(n)},\varnothing;0\bigr).
\end{align}
The index $n$ identifies a sample within the batch. Both passes use the same EMOTRANS denoiser $f_\Theta$, updating only its LoRA parameters, VAD encoder, and VAD injectors. $z_t^{(n)}$, $t_n$, $E_{\mathrm{text}}^{(n)}$, and $M_{\mathrm{text}}^{(n)}$ are the sample's noisy latent, timestep, text embeddings, and text mask; the subscript $t$ in $z_t^{(n)}$ takes the sampled value $t_n$. $Z_{\mathrm{vad}}^{(n)}$ is its VAD-token sequence, and $\widehat{w}_{t,n}$ its effective modulation weight. The operator $\pi_4$ retains only the first four noise- or velocity-prediction channels from PixArt's eight-channel output. The two $\widehat{y}$ terms denote the branch predictions. $\varnothing$ denotes absent VAD conditioning, and $0$ disables emotional residual injection. ADBO thus introduces neither a second network nor a separate set of content parameters.

The relative importance of the two supervision signals must also be determined across denoising stages. A fixed mixture ignores the greater need for content constraints at high noise levels and emotional refinement at low noise levels. ADBO therefore reuses TAVM's $\widehat{w}_t$ for sample-wise adaptive mixing. The per-sample losses are
\begin{equation}
\ell_{\mathrm{vad}}^{(n)}=\MSE\bigl(\widehat{y}_{\mathrm{vad}}^{(n)},y_t^{(n)}\bigr),\qquad
\ell_{\mathrm{text}}^{(n)}=\MSE\bigl(\widehat{y}_{\mathrm{text}}^{(n)},y_t^{(n)}\bigr).
\end{equation}
These are nonnegative scalar denoising losses, and $y_t^{(n)}$ is the shared target for the sample. MSE averages squared error over all channels and spatial elements. ADBO mixes the losses with the same effective weight used for VAD residuals:
\begin{align}
\ell_{\mathrm{ADBO}}^{(n)}
&=\sg(\widehat{w}_{t,n})\ell_{\mathrm{vad}}^{(n)}
+\bigl[1-\sg(\widehat{w}_{t,n})\bigr]\ell_{\mathrm{text}}^{(n)},\\
\mathcal{L}_{\mathrm{ADBO}}&=\frac{1}{B}\sum_{n=1}^{B}\ell_{\mathrm{ADBO}}^{(n)}.
\end{align}
The stop-gradient operator $\sg$ preserves its input in the forward pass and sets its backward gradient to zero. $\ell_{\mathrm{ADBO}}^{(n)}$ is the sample-wise adaptive dual-branch loss, and $\mathcal{L}_{\mathrm{ADBO}}$ averages it over the batch. Consequently, $\widehat{w}_{t,n}$ determines only the mixing ratio and receives no gradient through the weight-computation pathway.

\section{Experiments and Discussion}
\subsection{Experimental Setup and Implementation Details}
EMOTRANS uses pretrained PixArt-$\alpha$ as its base generator and is trained on EMOVAD. The base PixArt-$\alpha$ model, text encoder, and VAE remain frozen; only LoRA parameters, the VAD token encoder, and layer-specific VAD injectors are updated. Training uses a single NVIDIA A800 80 GB GPU, Python 3.10, AdamW, and half precision. Table~\ref{tab:setup} summarizes the environment and main hyperparameters.

We evaluate overall performance using CLIPScore, CLIP-IQA, and V/A/D-Error. CLIPScore measures similarity between the original content prompt and the generated image in CLIP representation space; higher values indicate better alignment of subjects, scenes, and semantic relationships with the text. Because EMOTRANS does not write VAD into the objective content prompt, this metric primarily assesses whether emotional modulation compromises semantic constraints, rather than directly measuring emotional accuracy. CLIP-IQA is a no-reference perceptual quality metric assessing clarity, naturalness, and overall visual acceptability; higher values indicate better perceptual image quality. Automatic emotion estimators may compress complex composition, metaphor, and artistic style into limited regression outputs. We therefore also conduct human evaluations using Ranking Consistency, human V/A/D-Error, Emotion Consistency, and Emotion Smoothness. Human error measures the absolute deviation between participants' estimates and target values. In Study II, Emotion Consistency and Emotion Smoothness use five-point Likert scales to assess whether an image group consistently conveys its target emotion and whether adjacent control states transition naturally, respectively. Higher scores are better.

\begin{table}[tbp]
\centering\small
\caption{Experimental environment and main hyperparameters.}\label{tab:setup}
\begin{tabular}{llll}
\toprule
Configuration & Value & Configuration & Value\\\midrule
Input resolution & $512\times512$ & LoRA rank & 4\\
Optimizer & AdamW & Weight decay & 0.01\\
Training epochs & 200 & Batch size & 64\\
Initial learning rate & $1\times10^{-4}$ & Baseline & PixArt-$\alpha$\\
Precision & fp16 & GPU & NVIDIA A800 80 GB\\
Target projections & q, k, v, out & Language & Python 3.10\\
VAD tokens & 4 & Condition dropout & 0.05\\
Fourier frequencies & 4 & $\lambda$ & 0.75\\\bottomrule
\end{tabular}
\end{table}

\subsection{Comparative Experiments}
The central premise of VCIM is to separate three-dimensional psychological coordinates from the content-text channel and inject them through an independent global conditioning pathway. To examine whether improvements stem from this structural separation, we compare PixArt-$\alpha$ without explicit emotion control, three common conditioning strategies, and a pipeline that uses GPT-4 to rewrite VAD as emotional text before generation with PixArt-$\alpha$.

Table~\ref{tab:comparison} reports emotional accuracy, content alignment, and image quality. All structural injection baselines and our method use the PixArt-$\alpha$ backbone. GPT-4 + PixArt-$\alpha$ first converts VAD into a prompt containing emotional descriptions, then generates an image with PixArt-$\alpha$.

\begin{table}[tbp]
\centering\small\setlength{\tabcolsep}{5pt}
\caption{Quantitative comparison of emotion-injection strategies. Bold indicates the best value in each column.}\label{tab:comparison}
\begin{tabular}{lrrrrr}
\toprule
Method & A-Error $\downarrow$ & V-Error $\downarrow$ & D-Error $\downarrow$ & CLIPScore $\uparrow$ & CLIP-IQA $\uparrow$\\\midrule
PixArt-$\alpha$ & 2.217 & 2.171 & 2.230 & \textbf{26.641} & 0.915\\
Cross Attention & 1.913 & 1.895 & 2.006 & 26.177 & \textbf{0.921}\\
Time Embedding & 1.931 & 1.870 & 1.965 & 26.476 & 0.848\\
Textual Inversion & 1.947 & 1.836 & 1.961 & 22.754 & 0.689\\
GPT-4 + PixArt-$\alpha$ & 1.891 & 1.614 & 1.801 & 25.925 & 0.905\\
Ours (EMOTRANS) & \textbf{1.792} & \textbf{1.506} & \textbf{1.719} & 25.812 & 0.903\\\bottomrule
\end{tabular}
\end{table}

EMOTRANS obtains the lowest error on all three dimensions. Relative to the original PixArt-$\alpha$, A-, V-, and D-Error decrease from 2.217, 2.171, and 2.230 to 1.792, 1.506, and 1.719, respectively. This indicates that objective text alone does not reliably specify emotional coordinates, whereas independent VAD conditions provide substantial control along all three psychological dimensions. Compared with the strongest prompt-rewriting baseline, GPT-4 + PixArt-$\alpha$, EMOTRANS further reduces the three errors by 5.2\%, 6.7\%, and 4.6\%, respectively. The results suggest that independent VAD tokens and layer-wise queries preserve continuous intensity and dimensional distinctions more effectively than compressing emotion into natural-language adjectives.

Cross Attention, Time Embedding, and Textual Inversion improve upon PixArt-$\alpha$, but do not jointly address numerical--visual representational mismatch, competition between content and emotion channels, and differences among denoising stages. In particular, Textual Inversion lowers CLIPScore and CLIP-IQA to 22.754 and 0.689, suggesting that emotional text tokens can disrupt the original semantic space. EMOTRANS achieves a CLIPScore of 25.812, only 0.113 below GPT-4 + PixArt-$\alpha$, and its CLIP-IQA is only 0.002 lower. Relative to PixArt-$\alpha$, these metrics decrease by approximately 3.1\% and 1.3\%. The model therefore trades limited changes in semantic alignment and perceptual quality for more accurate, consistent continuous emotion control across all three dimensions, providing a practical balance between content preservation and emotional controllability.

\begin{figure}[tbp]
\centering
\includegraphics[width=1\linewidth]{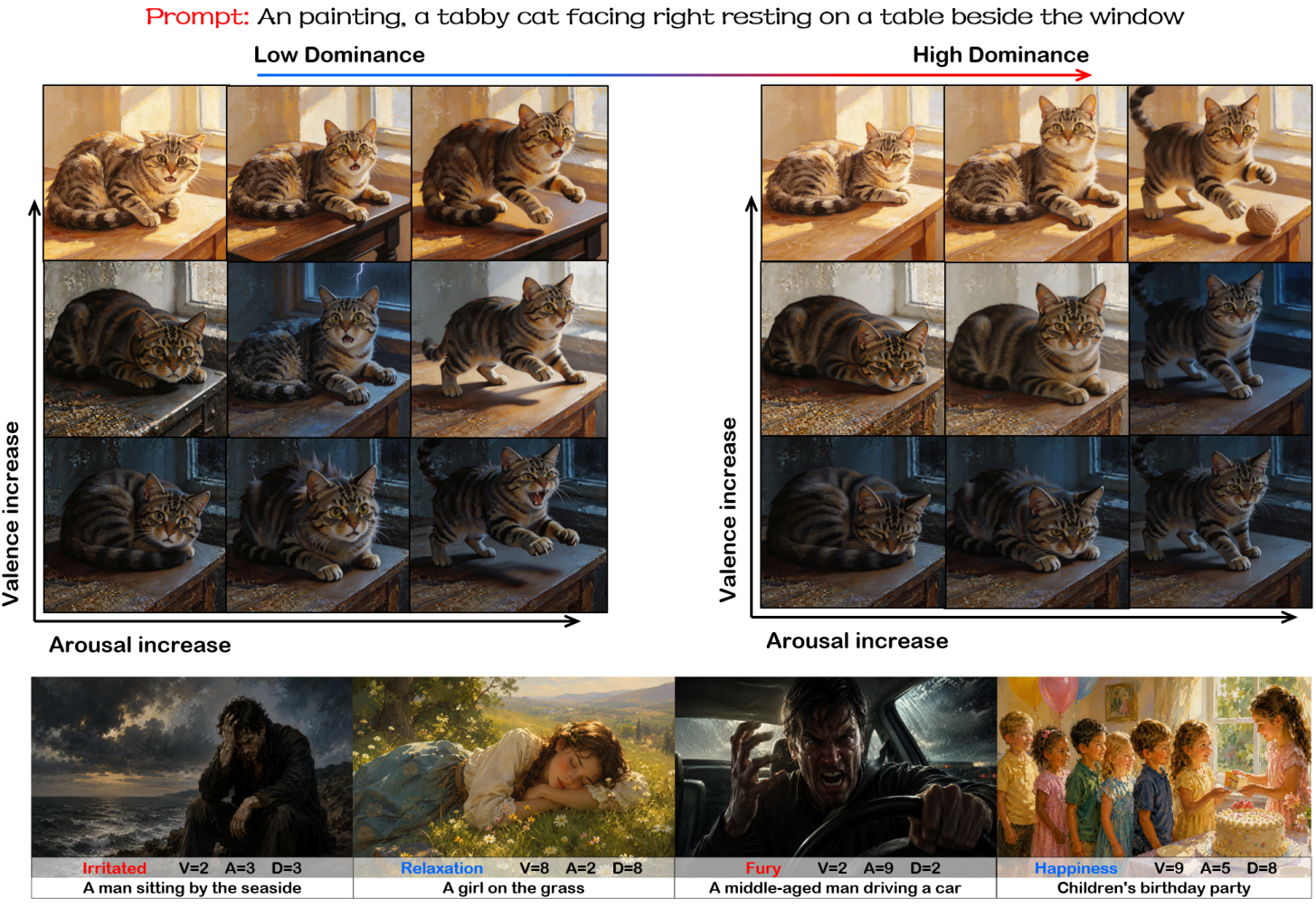}
\caption{Continuous VAD control and representative emotional expressions. The upper panels show V--A grids at low and high dominance for the same tabby-cat prompt. The lower panel shows four representative VAD combinations: Irritated, Relaxation, Fury, and Happiness.}
\label{fig:continuous}
\end{figure}

Lower average error alone cannot establish whether a model learns continuous, interpretable, and complementary dimensional control. Figure~\ref{fig:continuous} therefore varies V, A, and D under a fixed content condition and additionally examines representative emotional combinations. This tests whether three-dimensional coordinates map to coordinated visual changes rather than merely triggering discrete emotion templates.

In the upper tabby-cat grids, the subject category, tabletop scene beside a window, and overall painterly form remain recognizable across conditions, indicating that VAD modulation does not replace the content text. Increasing valence shifts the image from cold, dark, oppressive tones toward brighter, warmer illumination, with the cat appearing more relaxed. Increasing arousal changes the posture from curled-up stillness toward alertness or impending movement, accompanied by stronger local contrast and visual tension. Dominance is not a simple duplicate of valence or arousal. At high dominance, cats tend to show more upright or forward-facing postures, clearer gazes, and a stronger spatial presence. Low dominance more often produces contracted, tabletop-hugging, or passive body language. Thus, the three axes yield observable, nonidentical changes within the same semantic structure. D adds a sense of control and agency that is difficult to describe within the V--A plane alone.

The lower examples show corresponding differences. Irritated ($V=2,A=3,D=3$) uses low brightness and closed posture; Relaxation ($V=8,A=2,D=8$) shows soft sunlight, little movement, and an open environment; Fury ($V=2,A=9,D=2$) emphasizes high-arousal negative valence through tense movement and strong contrast; and Happiness ($V=9,A=5,D=8$) expresses positive energy through warm colors, interaction, and celebratory composition. These examples indicate that EMOTRANS jointly adjusts lighting, posture, composition, and atmosphere within a shared semantic structure, rather than reducing VAD to a single color filter.

\begin{figure}[tbp]
\centering
\includegraphics[width=1\linewidth]{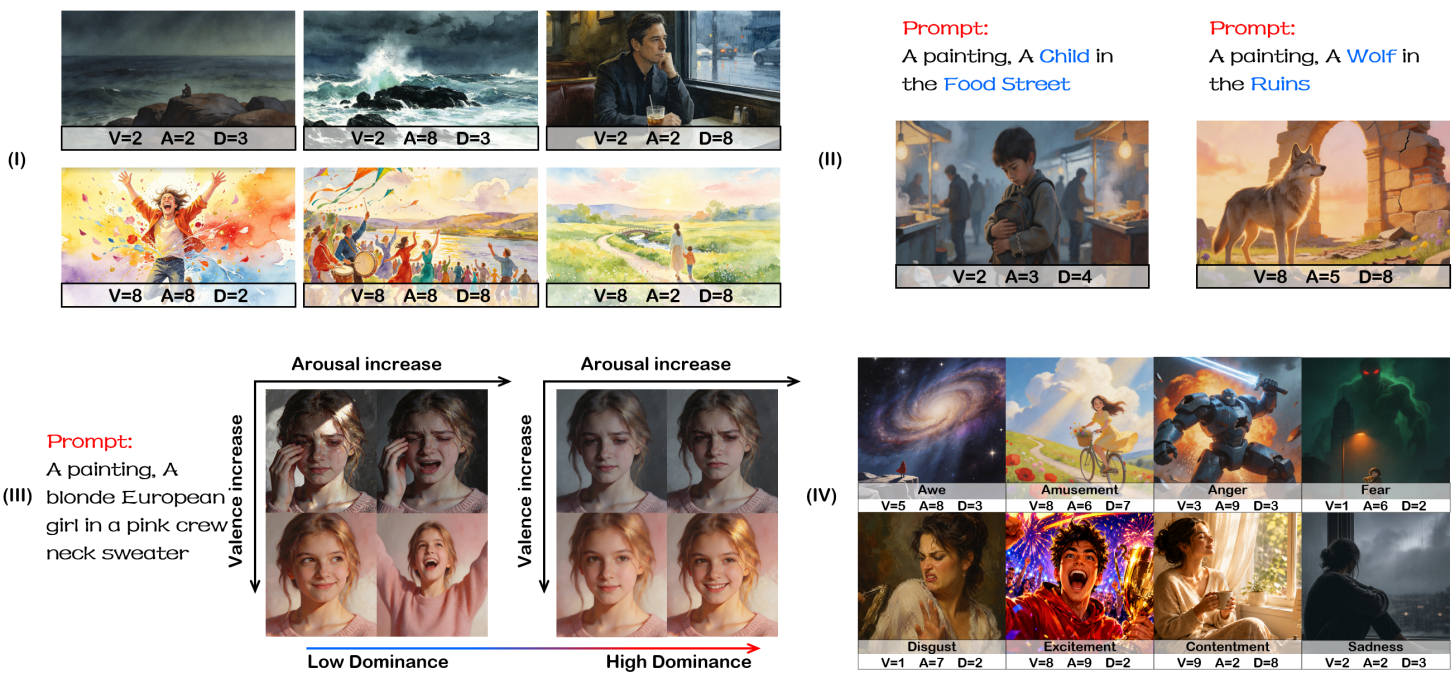}
\caption{Validation of VAD control mechanisms. (I) VAD-driven generation with a nearly empty prompt. (II) VAD conditions opposing the text's emotional prior. (III) Continuous control under a fixed objective character description. (IV) Cooperation between explicit emotion words and consistent VAD conditions.}
\label{fig:mechanism}
\end{figure}

An independent VAD pathway should remain effective with nearly empty prompts, conflicting textual priors, and aligned conditions. In Figure~\ref{fig:mechanism}(I), the prompt is only ``a painting.'' Low V produces dark seascapes, storms, or introspective figures, whereas high V shifts toward celebrations and bright landscapes. With V fixed, increasing A strengthens waves, movement, and color impact, while increasing D more strongly changes perceived agency, spatial openness, and scene organization. The examples suggest that VCIM's VAD tokens can directly drive visual emotion without explicit emotion words.

Figure~\ref{fig:mechanism}(II) examines whether VAD can override opposing textual priors. ``Food street'' commonly suggests a lively, pleasant setting, but $V=2,A=3,D=4$ yields low brightness, inward postures, and a quiet atmosphere. ``Wolf in the ruins'' tends to imply danger and desolation, whereas $V=8,A=5,D=8$ produces warm colors and a more positive spatial impression. Subjects and scenes remain stable, suggesting that the independent style pathway can alter textual emotional tendencies while retaining content semantics.

Figure~\ref{fig:mechanism}(III) illustrates fine-grained control with a fixed description of a blonde girl. V primarily shifts expression and illumination from negative to positive; A intensifies facial and bodily movement. At similar V--A positions, high D appears more stable, confident, and direct, while low D appears more vulnerable or passive. Identity, clothing, and composition are largely retained, concentrating changes on emotional expression. Figure~\ref{fig:mechanism}(IV) shows cooperation between explicit emotion words and consistent VAD: high-arousal Anger and Excitement have stronger movement and contrast; Contentment with low A, high V, and high D is calmer and brighter; and Fear or Sadness with low V and low D appears more oppressive or withdrawn. Text supplies discrete categories, while VAD adds continuous intensity and dimensional structure.

\begin{figure}[tbp]
\centering
\includegraphics[width=1\linewidth]{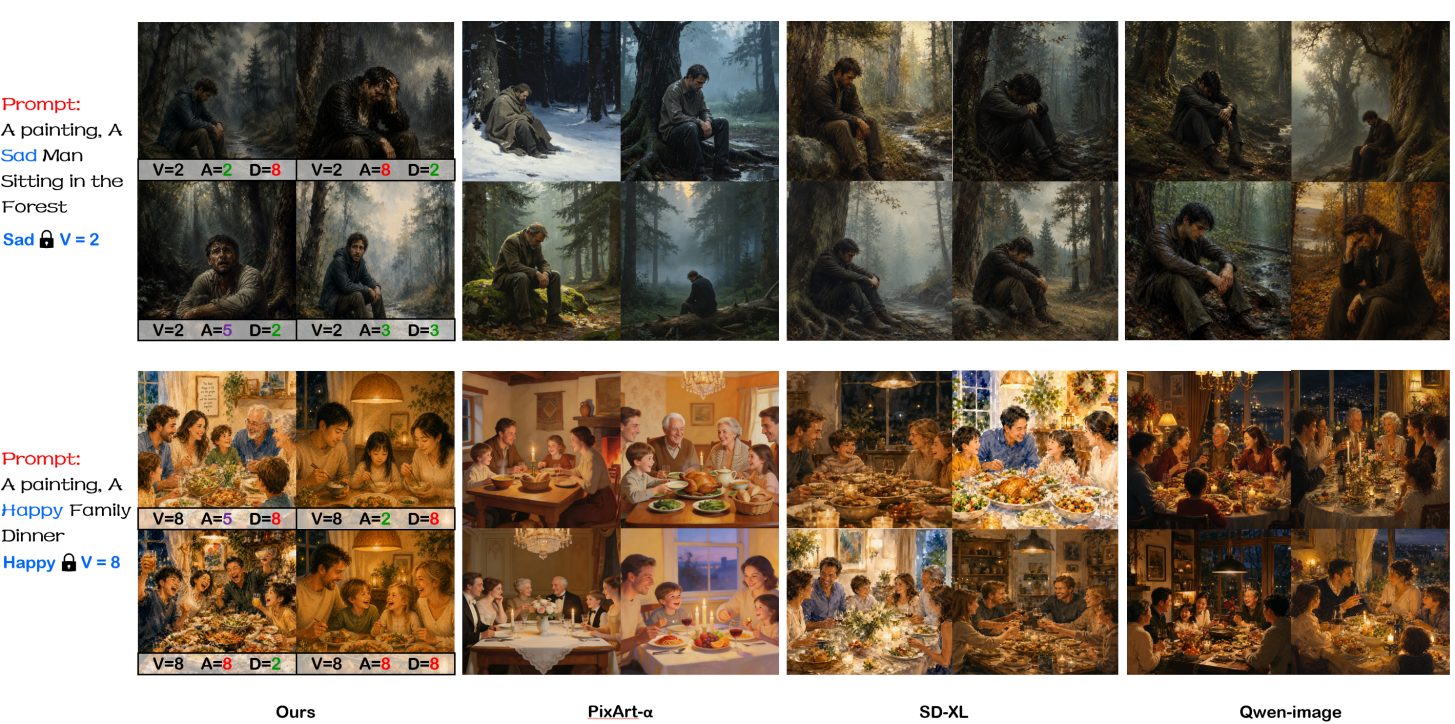}
\caption{Qualitative comparison of EMOTRANS, PixArt-$\alpha$, SD-XL, and Qwen-image. The upper example shows a sad figure in a forest; the lower example shows a happy family dinner. EMOTRANS varies intensity and dominance within an emotion using VAD, while the other models use textual emotional descriptions.}
\label{fig:models}
\end{figure}

Automatic metrics may overlook whether emotion is precisely expressed in the same content instance. We therefore visually compare EMOTRANS with other mainstream models in fixed scenarios. Figure~\ref{fig:models} uses explicitly negative Sad and positive Happy prompts. EMOTRANS injects numerical VAD into PixArt-$\alpha$, while the corresponding emotions are expressed as prompts for the general-purpose models, including SD-XL and Qwen-image. The four methods are shown side by side to examine visible differences in emotional expression and content preservation.

All models express the broad categories of sadness and happiness, indicating that discrete emotion words provide directional priors. EMOTRANS additionally adjusts arousal and dominance at fixed valence. In the sad example, low A favors stillness, contemplation, and enclosed composition, whereas high A increases tense posture and environmental motion. Changes in D further affect whether a figure appears pressured and withdrawn or retains agency. In the happy dinner example, low A resembles a gentle family reunion; high A produces more pronounced laughter, gestures, and group interaction. D alters the participants' initiative and centrality in the composition. General-purpose models produce visually strong, category-appropriate emotional images, but their samples resemble different versions of sadness or happiness without fine-grained variation explicitly organized by independent numerical axes.

\subsection{Ablation Studies}
\begin{table}[tbp]
\centering\small
\caption{Component ablation of VCIM, TAVM, and ADBO. Bold indicates the best value in each metric column.}\label{tab:ablation}
\begin{tabular}{cccrrrr}
\toprule
VCIM & TAVM & ADBO & A-Error $\downarrow$ & V-Error $\downarrow$ & D-Error $\downarrow$ & CLIPScore $\uparrow$\\\midrule
-- & -- & -- & 2.217 & 2.171 & 2.230 & 26.641\\
$\checkmark$ & -- & -- & 1.893 & 1.716 & 1.863 & 25.476\\
-- & $\checkmark$ & -- & 2.211 & 2.164 & 2.235 & 25.912\\
-- & -- & $\checkmark$ & 2.197 & 2.166 & 2.237 & 26.067\\
$\checkmark$ & $\checkmark$ & -- & 1.822 & 1.701 & 1.814 & 25.654\\
-- & $\checkmark$ & $\checkmark$ & 2.220 & 2.148 & 2.229 & \textbf{26.689}\\
$\checkmark$ & -- & $\checkmark$ & 1.815 & 1.718 & 1.871 & 25.810\\
$\checkmark$ & $\checkmark$ & $\checkmark$ & \textbf{1.792} & \textbf{1.506} & \textbf{1.719} & 25.812\\\bottomrule
\end{tabular}
\end{table}

We conduct ablations to evaluate the contributions of the three modules and the VAD injection mechanism. Table~\ref{tab:ablation} shows that removing VCIM increases A/V/D-Error from 1.792/\allowbreak{}1.506/\allowbreak{}1.719 to 2.220/\allowbreak{}2.148/\allowbreak{}2.229. Retaining only TAVM or ADBO produces similarly large errors. These modules regulate residual strength and dual-branch constraints, respectively; without VCIM to encode low-dimensional VAD into global tokens and establish independent cross-attention, they have no effective emotional representation to modulate. VCIM alone reduces errors to 1.893/\allowbreak{}1.716/\allowbreak{}1.863, supporting its foundational role.

Removing TAVM while retaining VCIM and ADBO increases the three errors by 0.023, 0.212, and 0.152, with almost unchanged CLIPScore. Fixed injection therefore primarily weakens control accuracy, particularly for V and D. TAVM suppresses emotional interference at high noise levels and strengthens lighting, color, and posture modulation at low noise levels, favoring content recovery before emotional refinement. Removing ADBO while retaining VCIM and TAVM increases errors to 1.822/\allowbreak{}1.701/\allowbreak{}1.814 and reduces CLIPScore from 25.812 to 25.654. This suggests that the text-only branch helps preserve subjects and scenes while discouraging content changes made to fit emotional targets. TAVM and ADBO reuse the same temporal weight, aligning feature injection and optimization. Although TAVM + ADBO without VCIM attains the highest CLIPScore, 26.689, its errors remain 2.220/\allowbreak{}2.148/\allowbreak{}2.229. Greater similarity to the original generator can therefore improve textual similarity without enabling continuous VAD control.

\begin{figure}[tbp]
\centering
\includegraphics[width=1\linewidth]{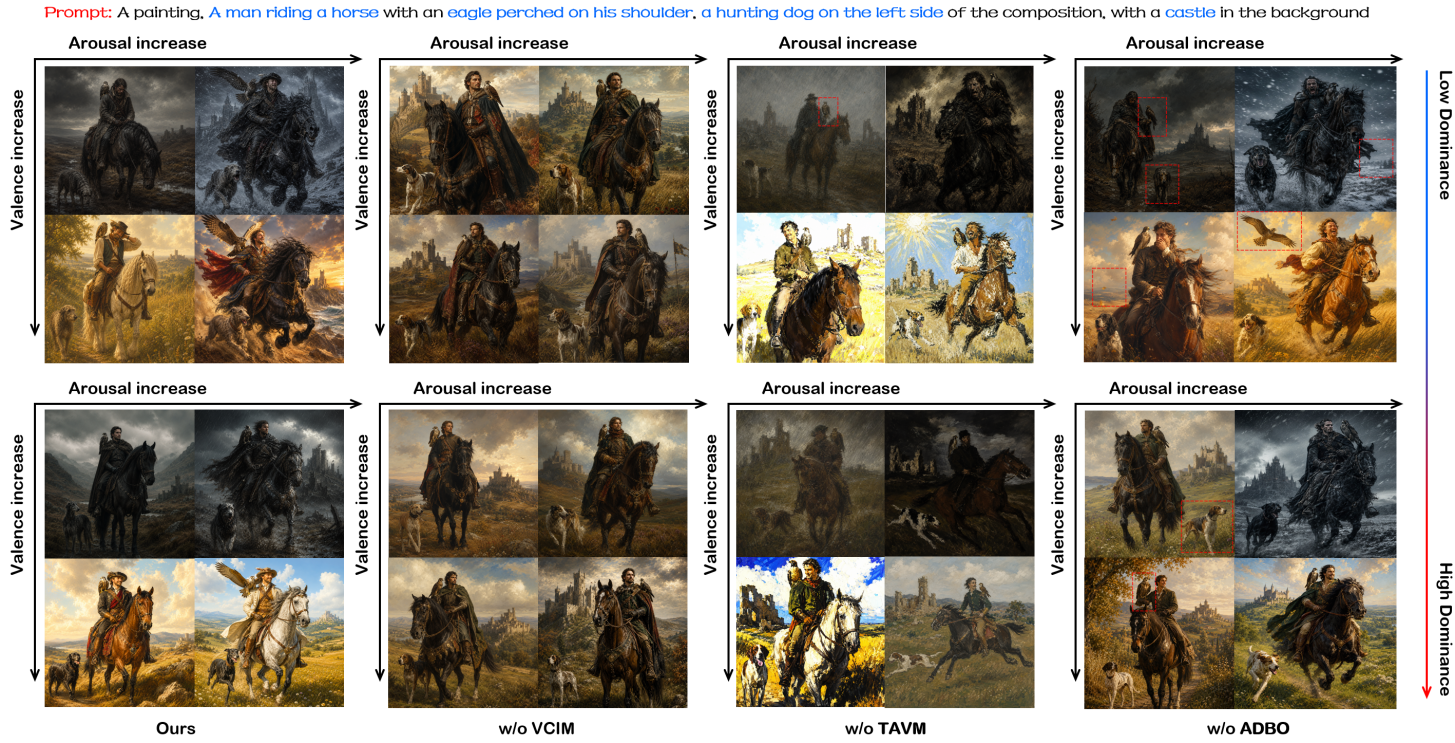}
\caption{Qualitative ablation of VCIM, TAVM, and ADBO. All variants use the same complex objective prompt. Red dashed boxes highlight missing or misplaced key content and semantic inconsistencies.}
\label{fig:ablation}
\end{figure}

Table~\ref{tab:ablation} quantifies overall trends but does not show how textual alignment deteriorates. Figure~\ref{fig:ablation} provides visual ablations using the same complex prompt: a man on horseback, an eagle on his shoulder, a hound on the left, and a castle in the background. Images are generated over VAD grids. Red dashed boxes identify prompt-inconsistent regions or missing key objects, relating each failure pattern to the corresponding module's role.

The full model retains the rider, horse, eagle, hound, and distant castle under low and high dominance and different V--A combinations. It expresses continuous emotion through weather, brightness, movement intensity, and postural agency. Without VCIM, subject content remains relatively stable, but emotional separation among grid positions weakens; color and atmosphere changes are conservative, with little continuous emotional variation. Without TAVM, some samples exhibit abrupt stylistic changes, overexposure, or structural instability, and the eagle, hound, and background relationships become less distinct. This suggests that a lack of suppression at high noise levels allows emotional residuals to compete prematurely with structural recovery. Without ADBO, the eagle, hound, or background objects highlighted by red boxes are more often missing, misplaced, or replaced, suggesting that the model may strengthen emotional appearance by rewriting content.

\subsection{Human Evaluation}
Emotion is a continuous psychological variable ultimately experienced through human perception, while automatic regressors can be affected by training domain, compositional bias, and artistic style. We therefore adopt EmotiCrafter's two-stage human-evaluation protocol~\cite{r12} and extend it from V--A to three-dimensional VAD including dominance. We examine whether participants recover the monotonic order of input VAD values and perceive continuous changes as emotionally consistent and smooth. The comparison uses GPT-4 + PixArt-$\alpha$, the prompt-rewriting baseline with the strongest three-dimensional emotion accuracy in our comparative experiments. We recruit 30 volunteers with different ages and disciplinary backgrounds through relevant professional communities and crowdsourcing platforms, and organize two evaluation stages.

\paragraph{Study I: Single-axis ranking and numerical estimation.}
We prepare three groups of single-axis image sequences for V, A, and D. Each group contains five images with target values 1, 3, 5, 7, and 9. Images are presented in random order. Participants first reorder them by perceived intensity and then estimate each image's V/A/D values. Ranking Consistency is computed using Kendall's $\tau$-b between the participant and target rankings~\cite{r12,r39}, while absolute error compares human estimates with input values. These measures assess monotonic controllability and absolute scale calibration, respectively.

\paragraph{Study II: Emotional consistency and smoothness.}
Participants view image groups with continuously varying VAD and rate Emotion Consistency and Emotion Smoothness on five-point Likert scales. To avoid the excessive viewing burden of a full three-dimensional grid, we use three orthogonal slices: a V--A grid at $D=5$, a V--D grid at $A=5$, and an A--D grid at $V=5$. Each comparison therefore varies only two axes while covering all three pairwise dimensional interactions.

\begin{table}[tbp]
\centering\small
\caption{Two-stage human evaluation (mean $\pm$ standard deviation). Bold indicates the better result between the two methods.}\label{tab:human}
\begin{tabular}{llcc}
\toprule
Study & Metric & EMOTRANS & GPT-4 + PixArt-$\alpha$\\\midrule
I & A-Ranking Consistency $\uparrow$ & $\mathbf{0.805\pm0.195}$ & $0.716\pm0.276$\\
I & V-Ranking Consistency $\uparrow$ & $\mathbf{0.852\pm0.207}$ & $0.792\pm0.248$\\
I & D-Ranking Consistency $\uparrow$ & $\mathbf{0.796\pm0.248}$ & $0.657\pm0.292$\\
I & A-Error $\downarrow$ & $\mathbf{1.109\pm0.992}$ & $1.417\pm1.176$\\
I & V-Error $\downarrow$ & $\mathbf{1.298\pm1.124}$ & $1.418\pm1.006$\\
I & D-Error $\downarrow$ & $\mathbf{1.208\pm1.006}$ & $1.465\pm1.141$\\
II & Emotion Consistency $\uparrow$ & $\mathbf{4.261\pm0.869}$ & $3.860\pm0.884$\\
II & Emotion Smoothness $\uparrow$ & $\mathbf{4.174\pm0.801}$ & $3.720\pm0.905$\\\bottomrule
\end{tabular}
\end{table}

As shown in Table~\ref{tab:human}, EMOTRANS achieves A/V/D-Ranking Consistency of 0.805/\allowbreak{}0.852/\allowbreak{}0.796, exceeding GPT-4 + PixArt-$\alpha$'s 0.716/\allowbreak{}0.792/\allowbreak{}0.657 by 0.089/\allowbreak{}0.060/\allowbreak{}0.139. The largest gain occurs on D, suggesting that participants more reliably distinguish increasing control, whereas prompt rewriting more readily mixes dominance with V--A semantics. Numerical estimates follow the same pattern: EMOTRANS obtains A/V/D-Error of 1.109/\allowbreak{}1.298/\allowbreak{}1.208, below the baseline's 1.417/\allowbreak{}1.418/\allowbreak{}1.465 by 0.308/\allowbreak{}0.120/\allowbreak{}0.257. The joint improvement in dominance ranking and absolute calibration on the 1--9 scale suggests that the model organizes gaze, posture, and spatial cues into a quantifiable dominance response.

In Study II, Emotion Consistency and Emotion Smoothness reach 4.261 and 4.174, improving over 3.860 and 3.720 by 0.401 and 0.454. The standard deviation of Smoothness also decreases from 0.905 to 0.801. These results suggest that independent VAD injection preserves emotional direction and produces smooth transitions, consistent with the motivations for TAVM's progressive residual modulation and ADBO's preservation of content structure.

\subsection{Failure Cases and Limitations}
Average performance describes overall behavior over the main data distribution, but does not fully characterize the tails of the conditioning space or responses to mutually contradictory constraints. It is therefore necessary to identify inputs on which EMOTRANS is more likely to fail, and to distinguish model limitations from intrinsically unsatisfiable conditions. We find that failures mainly occur at extreme VAD coordinates, particularly when one or more dimensions approach the endpoints of the scale. These combinations are generally less represented in real art samples than intermediate regions, requiring extrapolation of strong emotional styles from limited visual evidence. As target intensity increases, color, brightness contrast, brushwork rhythm, or postural tension may saturate, and local textures and small structures may be slightly affected. In some samples, cues associated with one dimension become too strong and weaken smooth transitions between neighboring VAD settings. This behavior is more consistent with extrapolation errors in distribution tails than with random instability across the ordinary VAD range.

Another limitation is that more stable three-dimensional emotion control comes at a small cost in image quality. The objectives are not identical: the base model primarily fits the text-conditioned visual distribution, whereas EMOTRANS must additionally use limited representational capacity to respond to independent VAD conditions and intentionally adjust color, lighting, texture, and compositional tension, all of which are closely related to perceptual quality metrics. We interpret the observed difference as a Pareto trade-off among content fidelity, perceptual quality, and emotional controllability, rather than evidence of general degradation. The model exchanges controlled, limited quality variation for interpretable emotional responses that can be adjusted along individual axes.

\section{Conclusion and Future Work}
We introduce EMOTRANS, an emotion-driven text-to-image model that treats VAD psychological coordinates as interpretable, continuous emotional control variables independent of textual semantics. VCIM, TAVM, and ADBO coordinate conditional representation, diffusion timing, and optimization, while EMOVAD provides separate supervision through paintings, objective descriptions, and human VAD ratings. Objective and human evaluations show that EMOTRANS reduces control errors along all three psychological dimensions and produces continuous responses that observers can reliably order. In particular, dominance captures relationships involving control, pressure, and subordination that are difficult to describe within the V--A plane. The experiments demonstrate a favorable balance between emotional controllability and image quality.

Future work will extend the VAD representation, annotation, and injection mechanisms established in the artistic domain to general-purpose AI image generation. We will investigate cross-domain scale calibration and emotional-cue transfer in natural images, content spanning multiple styles, and complex semantic scenes, aiming to maintain comparable meanings for the same psychological coordinates across visual domains. We will also address the boundary cases identified here: augmenting distribution tails and modeling uncertainty to improve robustness at extreme VAD values; incorporating text--emotion conflict detection and confidence-aware dynamic gating to coordinate semantics and emotion under strong conflicts; and using more refined multi-objective optimization to reduce the image-quality cost of emotional control. The ultimate goal is to develop emotion from prompt modification that depends on wording experience into an interpretable, calibratable generative interface reusable across domains.

\end{document}